# A Study of Scaling Issues in Bayesian Belief Networks for Ship Classification

S. A. Musman*
Artificial Intelligence Technical Center.
The MITRE Corporation
McLean, Va, 22102

L. W. Chang
Code 5510
Naval Research Laboratory
Washington DC, 20375

## Abstract

The problems associated with scaling involve active and challenging research topics in the area of artificial intelligence. The purpose is to solve real world problems by means of AI technologies, in cases where the complexity of representation of the real world problem is potentially combinatorial. In this paper, we present a novel approach to cope with the scaling issues in Bayesian belief networks for ship classification. The proposed approach divides the conceptual model of a complex ship classification problem into a set of small modules that work together to solve the classification problem while preserving the functionality of the original model. The possible ways of explaining sensor returns (e.g., the evidence) for some features, such as portholes along the length of a ship, are sometimes combinatorial. Thus, using an exhaustive approach, which entails the enumeration of all possible explanations, is impractical for larger problems. We present a network structure (referred to as Sequential Decomposition, SD) in which each observation is associated with a set of legitimate outcomes which are consistent with the explanation of each observed piece of evidence. The results show that the SD approach allows one to represent feature-observation relations in a manageable way and achieve the same explanatory power as an exhaustive approach.

## 1 Introduction

In the domain of ship classification there are potentially hundreds of candidate targets that can be observed. In the past several pilot studies [Booker 1988][Musman, Chang & Booker 1993] have demonstrated the feasibility and applicability of using Bayesian belief networks to solve the ship classification problem. However, that work only demonstrated small examples of how the problem can be solved. In both of the above cases the networks compared only 10-12 target types.

Because of the large number of targets that are present in the ship classification problem, there are potential difficulties and pitfalls which exist when trying to scale up the example networks shown in the pilot studies to create a system which is capable of classifying the thousands of ship targets present in the world. At the moment there are more than 640 military combatant ship classes in the world and there are over 10,000 types of Commercial and Auxiliary craft. In our work we have focused on the task of identifying only the combatant targets.

In order to be able to address such a large class problem we use the same coarse to fine hierarchical classification techniques described by [Clancy 1984] [Chandrasekaran 1986] by defining a taxonomy for the ship classification problem. In addition to this, because of the potential complexity problems which exist in creating belief networks (each associated with different levels of the hierarchy), we must also ensure that the internal structure of each network is properly designed so that it can address the scaling issues normally associated with the addition of new targets or features. Examples of these two issues will be given.

## 2 Overview

Ship classification involves the use of more than 50 features to differentiate between target classes. As with many other classification problems, often several target classes are very similar in appearance and additionally there can be a substantial variation of each individual target's specifics within the same target class. This latter characteristic of the problem is normally caused by making structural modifications or the addition of new weaponry after a ship has been deployed.

Although there are several alternative ways in which to decompose the large problem into a hierarchical solution, we have endeavored to perform this operation in a manner that creates partial conclusions that have an intuitive meaning to the analyst. While it is possible (and is sometimes appropriate) to create sub-conclusions in our taxonomy that are defined by the separability of the features, it makes much more sense to relate the taxonomy categories to information about the target that is indicative of its mission or military capability. Figure 1 demonstrates the hierarchical breakdown of the ship classification problem from detection of a target on a radar PPI down to the target's naval class designation.

---

* Scott Musman used to be with NRL, Code 5362

At the upper levels of our taxonomy we normally assign priors which reflect the fact that each possible target class is equally likely. This allows our problem solving to detect the target which best explains the observed evidence. Such priors are not absolute and are not intended to indicate the actual frequency of occurrence of each target type in the world. Additionally, our priors are not considered to be firm and fixed. Instead they are expected to be tailored to match each specific mission scenario based on the effects of prior intelligence and other activity associated with the area of the world being analyzed (i.e., in the Mediterranean you are less likely to come across a Chinese ship, or a specific class of targets may be known to be in port and under repair). These "prior" values are expected to be calculated using a separate belief network that is designed to fuse such information from diverse intelligence reports.

Belief networks are associated with each level of the taxonomy hierarchy. Each network uses appropriate features to attempt to differentiate between the hypotheses at that level. If the observed evidence can differentiate between the candidate hypotheses at the given level of the taxonomy then the problem solving continues by loading a network associated with the next more specific level of the taxonomy. But, if the evidence yields an inconclusive result then the problem solving is suspended and the most specific result obtained from the taxonomy is returned to the user. This approach is described in more detail in [Musman, Chang & Booker 1993].

As a result of having many different networks associated with solving the problem, it is necessary to construct the networks to be *modular*. This will allow us to use the results from one network as a single piece of evidence in another network, or will allow us to use the results of one network as priors in another network. This approach has advantages and disadvantages. An advantage of this approach is that the networks loaded into memory tend to be smaller and simpler than networks which address the whole problem. As a result it is possible to use dynamically computed measures of informativeness to efficiently order the acquisition of evidence [Pearl 1989][Musman, Chang & Booker 1993]. The primary disadvantage of this modular approach is that the use of such informativeness measures is restricted to the single smaller networks. Thus, the ability to change the focus of attention is much more limited than it could be. There is a trade-off in flexibility vs computation, that must be considered when creating the network modules.

As a result of the large number of targets which must be modeled by our belief networks, if one were to build a network (or network modules) which compare each target to each other target type, there would be a large amount of duplicate structure within the network associated with the fact that the targets are built of essentially similar components (i.e. all targets have superstructure mounted on the deck, etc.). This means that every time a target is added or deleted from such a network it is necessary to add or delete all of the supporting *structural* links associated with that target. This makes such a network much harder to maintain (Figure 2).

To circumvent the above problem, we have chosen to utilize the fact that all ship targets have essentially the same structural makeup. Thus, the main difference between the targets is found in the details about how the features appear on the targets' components, and is not due to the fact that some targets have different components than others (i.e., all targets have both superstructure and masts but the shapes of these features are different). By using this constraint, it is possible to create a single network which represents the structural makeup of a target (Figure 3). We then use this network by adding evidence which indicates the errors between the observed evidence and a single specific target.

In order to use the network shown in Figure 3 it is necessary to re-instantiate the network for each known target type by adding evidence which represents a measure of the error between the observed evidence and the expected description for each specific target. Thus, evidence is added to the network comparing the observations to target1, the result is obtained, and then the process is repeated for each known target type. This network has two top level hypotheses: Target and Other. The "Target" hypothesis represents a known target type (which is perfectly described when there is no error between the observed evidence and the expected description). The "Other" hypothesis represents randomness where all of the errors are equally likely. A {T | O} network will be referred to here as $T_i$ -module, for $i$'th ship class.

The advantage of using this approach to solving the ship classification problem is that we now obtain two pieces of information about each type of target: We obtain a measure of how well the evidence matches the specific target type. This means that we can look at the final belief of the network to understand if the specific target being observed has been modified, or is somehow different from the prototypical example stored in our target database (i.e., this makes it possible to identify an unknown ship). It is still possible to compare the beliefs in target1 vs target2, etc., to produce the probability of {target1,target2,....targetN} exactly as would have been calculated by the network shown in Figure 2.

We will describe more details about using this approach later in the paper.

## 3 Network Structure

Under most conditions it is appropriate to use the belief at each node to compare candidate targets. There are however, certain conditions which can cause the value of belief to be other than what is desired. This condition occurs when there can be ambiguity associated with a single observation. When such ambiguity exists, the target hypothesis that contains the largest number of possible alternatives which may match the observation will be assigned the highest belief. While this assignment of belief is correct, it is often found that the analyst is assured that the single observation can only match one of the possible outcomes (i.e., by being confident that he can identify the observation as being a single one and definitely not two coincident observations that may appear as one). When this happens, the result the analyst is expecting to see is one which compares only the single best possible explanation for each hypothesis versus the best explanation for each other hypothesis. In Bayesian belief networks this comparison of best explanations is called bel* and is defined in [Pearl 1989] as follows: $bel^*(x) = \max_{w_x} p(x, w_x | e)$
where x stands for a variable, e stands for pieces of evidence and $w_x$ stands for the instantiation of all variables, except x, on the belief network. Therefore, to find the best explanation is to search for the variables' instantiations that maximize probability distributions.

As an example of this phenomenon we propose the following example:

*It is often possible to identify ship targets at night by noting the number and location of portholes along the length of the target. Each porthole location is noted as a percentage location along the length of the target where the bow is 0% and the stern represents 100%. A typical measurement can be made in 10% intervals.*

This type of problem is hard to model using Bayesian belief networks because there are multiple causes for being able to observe a porthole along the length of the ship. Not only must a porthole be present at a specific location on the target for it to be observable, but also the lights inside of the porthole must also be illuminated. This means that for a given target the first observed porthole location on any given night may the first, second, or third porthole present on that target, and so on for the remaining portholes.

The easiest way in which to model this problem is to relate the number of lights observed with the number of lights on the target, and then exhaustively list out the possible permutations for how the lights on the target may be illuminated. This has the combinatorial behavior we mentioned earlier (Figure 4).

The network used to solve the problem shown in Figure 4 is interesting because it demonstrates the behavior and contrast of bel (i.e., the posterior probability evaluated from Bayesian belief network) and bel* for a given network node. If there are three targets, with the first having two portholes at 20% and 40% the ships length, the second having a single porthole at 40% and the third having two portholes at 50% and 70%. When given an observation located at 30% along the ships length, it is possible that this one observation can be caused by either the 1st or 2nd porthole on target-1, but could only be caused by the 1st porthole on target-2. No porthole on target-3 can possibly match this observed porthole. If this single piece of evidence is entered to the network the resultant belief at the Target node will correctly reflect the fact that target-1 is two times more likely than target-2 and Target-3 is discounted altogether (i.e., bel={0.66,0.33,0.0}). By contrast, the value of bel* reflects the fact that only one of the alternatives for target-1 can possibly match the single observation. This causes the bel* result to equally distribute its belief between target-1 and target-2 (i.e. bel*={0.5, 0.5, 0.0}).

To make this example a little more interesting we now add an additional constraint. It states that at night from a long distance it is often possible to confuse an open deck hatch for a porthole. If this happens, then it becomes possible to have the very first light observation actually be caused by an open deck hatch rather than the 1st, 2nd or 3rd porthole on the target. To make this problem tractable we limit our example to only allow the observation of one incorrect detection (i.e., we will only allow one deck hatch to be observed).

*As an additional constraint to the above problem, we will now allow the observation of a single incorrect porthole location (i.e., we assume that an incorrect observation is an open deck hatch) without wanting to penalize our final belief. This means that a single correct observation should yield the same resultant belief as observing one correct observation and one incorrect observation. If we observe two incorrect features, then this can be considered to be a non-coincidental error and we will expect the resultant belief to exclude any target which has more than one incorrect observation.*

Figure 5 demonstrates a simple network that produces the desired result when the bel* value of the top level node is queried. It is deigned to allow up to three observations but allow one of them to be incorrect (i.e., not match anything on a target) without penalizing the bel* of that target. It is worthwhile to examine the bel* value response to the piece of evidence shown in Figure 6. While the results demonstrate that the network appears to return the desired results, it is necessary to re-examine the structural relationships within this network to understand its scaling characteristics.

For a simple demonstration of this, take the problem shown above as an example. Let the possible observations of porthole-1, porthole-2, porthole-3 and a phony object (e.g., hatch) be denoted as 1, 2, 3 and W, respectively. Exhaustively listing all of the possible outcomes yields 23 of them:

1,2,3,W,

12,13,23,1W,2W,3W,W1,W2,W3

123,12W,13W,23W,W12,W13,W23,1W2,1W3,2W3

where one false detection of phony objects is allowed. This number is on the order of n! (where n is the number of features plus the number of false alarms). If we were to build a similarly structured network to solve a problem which would allow 6 observations and 2 false detections (which is more commensurate with real world conditions) then we would need to exhaustively list out 846 possible outcomes! This becomes impractical. The calculation of the number of possible outcomes is given in Appendix.

To overcome this problem we have designed a different network structure that is intended to produce the same bel* as the above network, but without producing the scaling characteristic noted above. We have called the approach Sequential Decomposition (SD) (Figure 7). It works by imposing a different set of independence assumptions about the observations than the above exhaustive approach. In this case the SD approach associates each observation only with legitimate outcomes. SD imposes on evidence from subsequent observations the constraints obtained from understanding the preliminary reasoning about evidence for the first few observations. That is, constraints are explicitly represented in SD structure. As a result of this, for the 3-porthole, 1-hatch problem we will have at most only seven possible ways of explaining observed pieces of evidence. The seven possible outcomes are {W1, 2, W2, 3, W, NO, O} after two observations, where NO denotes a constraint violation which can only be resolved by having this observation "not observed" (NO), and O means all other ship classes. The conditional probability of an evidence node, for example, $O_1$ given 2 has the same value as $O_1$ given W2, because the "W" in W2 simply means the violation of constraint. is violated. Also, because of the meanings of O and W, conditional probabilities of evidence nodes given O and W are assumed to be equally distributed. Note that 3 stands for 13 and 23. Since bel* selects the best instantiation, it's possible to compress multiple outcomes into one outcome and preserve bel*. This property of bel* leads to the equivalence relationship between the exhaustive and SD networks representations:

**Property 1.** The values of Bel*( $T_i$ ) (Bel*(O)) computed from Exhaustive and SD networks are equal.

That is, SD has desirable scaling properties for the computation of bel*. If we were to build a network to solve the 6 porthole two hatch problem then we would only need 16 hypotheses for each node. This number is much better than the 846 hypotheses required in the exhaustive approach.

While this new network is designed to produce the same value of bel* for any given set of observations (Figure 8), it is important to note that the bel values for the 2 networks are very different. This is because the independence assumptions for the evidence in each network are different. Because of this, different ambiguities exist in the different networks and it is these different ambiguities that cause the bel values to differ. In the case if the 3W problem described above, the difference between the exhaustive solution, and the SD solution is due to the fact that the exhaustive solution has a single node which explicitly states all of the network constraints and can indicate exactly which observations match or violate the implicit constraints of the network. In contrast to this, the SD network deals with the constraints only on an observation by observation basis. Thus while a belief in the hypothesis "W3" for O3-Alts implies that either observation 1 or observation 2 is assumed incorrect, unlike the exhaustive network, it does not state which of the observations is incorrect.

## 4 Integration of Belief Values

Our proposed approach to scaling is designed to work by instantiating a single network which models only the structure of each target and utilizes evidence in the form of an error measure. A separate bel* value is obtained for each target. The main advantage of using this approach is that it is very easy to add or delete targets to a classification system using this network because the network remains unaltered. The evidence added to this network is in the form of an error between observation and specific target, and these error measures are computed by comparing the observed evidence to feature values stored in a database. This means that simple adding or deleting database entries for targets is sufficient for altering the number of targets in the system.

Computationally, because we really wish to enter evidence to this network in a form which compares the probability that the observation is porthole-1, porthole-2, etc., on each target, in our work we have created functions which compute these likelihoods by comparing the observations to the database values. In doing this we have lost some of the characteristic benefits of using bi-directional inferences but have gained a substantial computational improvement.

In addition to the scaling advantages associated with this technique, it is easier to understand and analyze the behavior of the network. This is because we explicitly model the errors associated with each distinct feature type. These error values are always compared with a random distribution (our "Other" hypothesis) and it is thus much easier to ensure that one feature type (or evi-dence source) does not carry more weight in the decision making process than another feature. This characteristic can often be a significant problem when a system combines evidence from a variety of different and diverse sources.

Given a set of ship classes $T_1, ....., T_n$, the final decision for ship classification is based on integration of the results obtained from each individual module. Recall that, from earlier discussion, our decision is based only on observed evidence. Therefore, priors of $T_i$ and O are assumed to be equal (though this technique can easily deal with unequal priors) in all $T_i$-modules. In a $T_i$-module, let the proportion of $bel^*(T_i) : bel^*(O)$ be denoted by $r_i$ The final decision of $T_i$ is determined by comparing those $r_i$ s. In fact, if $bel^*(O)$ remains unchanged in different $T_i$-modules, then the ratio of $r_1 : ......r_n$ is simply $bel^*(T_1) : ...... : bel^*(T_n)$, which is exactly the ratio without using the {T | O} network model (i.e., all target classes are in one node). We describe this result in the following Property:

**Property 2.** The ratio of bel* of target classes, $bel^*(T_1) : ...... : bel^*(T_n)$, computed with using the {T | O} network model is equal to that computed without using the {T | O} network model. (Proof is given in [Musman & Chang 1993].)

Hypotheses can be rejected if there is no strong supporting evidence for them. This fact can be manifested from the ratio of bel*'s between $T_i$ and O. That is, if the ratios of $T_i$-modules are smaller than 1, for all i, then a statement such as "Target is something else." can be concluded.

When several features are evaluated, the method to calculate bel* is carried out by direct multiplication. The evaluation process is a recursive procedure which evaluates each $T_i$-module in turn.

In our system many of the conditional probability links contain subjective estimates of actual probability distributions. These distributions are based on both our analysis of the results of a limited training cycle with real data and our own extrapolations about how the limited training results may extend to the rest of the problem domain. We thus encourage a hybrid data-driven and model-based approach to estimating the conditional probability links.

When estimating our conditional probabilities we restrict our estimation processes to the comparison of likelihoods for each possible hypothesis. This allows us the opportunity to better compare the impact of evidence applied to each different hypothesis and thus allows us to compute the better balance of evidential weight noted above.

## 5   Discussion

A detailed analysis of the performance of our approach has not been possible. Although many of the techniques described above have been implemented in a prototype system, this prototype system is intended to perform completely automatic classification of ship images [Musman, Kerr, Bachmann 1993]. The design goal of the automatic recognition system was to classify the ship targets by using the same characteristic features that are used by human interpreters. This approach was taken so that partial classification results can still be used by image analysts as a decision aid. As a result of our design goal we have used human interpreters as an initial model for the uncertainty associated with the extraction of target features. Not surprisingly, it has been found on many occasions that the automatic feature extraction procedures do not behave in the same manner as the human interpreters. This characteristic introduces additional errors into our system. As more data on the actual behavior of the automatic feature extraction process is collected, the belief models can be updated to integrate the actual characteristics of the automatic feature extraction, rather than our original estimates.

Despite the problem of incorrectly modeling belief values for some of the the automatic feature extractors, the belief networks in the current prototype system have performed surprisingly well. On one occasion the results of the system did not match the ground truth information found in a log book which was supposed to indicate to us the correct target. After a detailed examination of the data, it was found that our classification system was correct and that the result found in the log was in error.

One another occasion the system was able to come up with the correct target as being noticeably more certain than the other possible candidate targets still under consideration after descending through the classification taxonomy. Even so, as a result of the two pieces of information we obtain about the observed evidence (i.e. the belief of a known target type versus other known target types; and the belief of the known targets versus randomness as specified by the $T_i$ modules) it was apparent that the observed evidence did not exactly match the best of our known target types. The ratio of T / O was noticeably smaller than it would have been if we had correctly described the target in our database, and the data quality was sufficiently good as to lead us to expect a better result. After some additional inspection of the test data, it was found that the image had been correctly identified but the ship under inspection had been modified by removing one of it's masts.

## 6 Conclusion

As with the previous studies, we have only been able to focus on a small portion of the ship classification problem. By combining the various techniques described separately in this and the previous papers, it is possible to create a target classification system which has the characteristics required for the ship classification problem.

The combined techniques have been tested in a prototype system which performed ship classification using approximately 15 features, for over 200 targets. While using 15 features wasn't normally enough to reliably accomplish complete classification, the final target ranking based on the likelihood of bel* was very useful as a decision aid.

**Appendix :** *Complexity of the 6/2 Problem*

The number of possible outcomes for 6 observations with tolerance of 2 false detections is 846. Let W denote the false detection. This value is obtained as follows:

Case 1. #(W)=0.

If there is no false detection, the number of outcomes is 63, i.e.,

$$C^6_1 + C^6_2 + C^6_3 + C^6_4 + C^6_5 + C^6_6.$$

Case 2. #(W)=1.

The outcomes for one false detection are equal to 249, i.e.,

$$1 + 2 \prod C^6_1 + 3 \prod C^6_2 + 4 \prod C^6_3 + 5 \prod C^6_4 + 6 \prod C^6_5.$$

Case 3. #(W)=2. In the case of two false detections, the number is 534, i.e.,

$$1 + (C^2_2 + 2) \prod C^6_1 + (C^3_2 + 3) \prod C^6_2 + (C^4_2 + 4) \prod C^6_3 + (C^5_2 + 5) \prod C^6_4.$$

Summation of the three values yields 846.

## References

Booker, L. (1988) "Plausible Reasoning in Classification Problem Solving." in *Image Understanding in Unstructured Environments* S. Chen (Ed) World Scientific Publishing, 1988.

Chandrasekaran, B. (1986) "Generic Tasks in Knowledge-based Reasoning: High-level Building Blocks for Expert System Design." *IEEE Expert*, Fall 1986.

Clancy, W. (1984) "Classification Problem Solving." Proceedings of the *Fourth National Conference on Artificial Intelligence*, Austin, Tx.

Musman, S., Chang L. & Booker L. (1993) "A Application of a Real-Time Control Strategy for Bayesian Belief Networks to Ship Classification Problem Solving." *J. of Pattern recognition and Artificial Intelligence*, to appear.

Musman S. & Chang L. (1992) "A Case Study of the Scaling Problem for Ship Classification." *NRL Memo Report*.

Pearl, J. (1989) *Probabilistic Reasoning in Intelligent Systems.* Morgan-Kauffman, Palo Alto, CA.

Musman, S., Kerr D., Bachmann, C. (1993) "Techniques for the Automatic Classification of ISAR Images" *Submitted to the IEEE Transactions on Aerospace Engineering* in review.

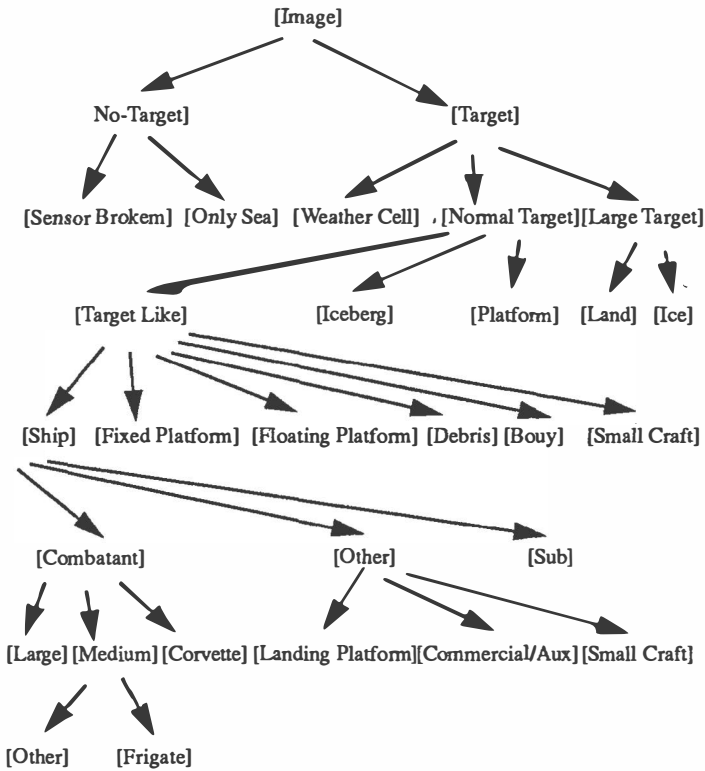

Figure 1: Ship Classification Heierarchy

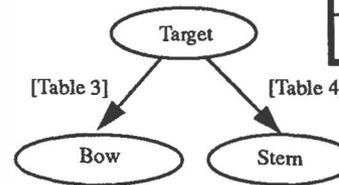

Table 3: Target -> Bow

|  | S1 | Other |
|---|---|---|
| <25% | 1.0 | 0.5 |
| >=25% | 0.0 | 0.5 |

Table 4: Target->Stern

|  | S1 | Other |
|---|---|---|
| Round | 0.7 | 0.33 |
| Curved | 0.3 | 0.33 |
| Straight | 0.0 | 0.33 |

Figure 3: A {T|O} network for the same features as in Fig 2.

Table 1: Target -> Bow

|  | S1 | S2 | S3 |
|---|---|---|---|
| <25% | 1.0 | 0.2 | 0.0 |
| >=25% | 0.0 | 0.8 | 1.0 |

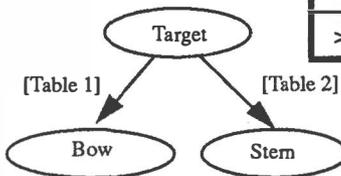

Table 2: Target->Stern

|  | S1 | S2 | S3 |
|---|---|---|---|
| Round | 0.7 | 0.1 | 0.2 |
| Curved | 0.3 | 0.8 | 0.1 |
| Straight | 0.0 | 0.1 | 0.7 |

Figure 2: A simple network with 3 targets and 2 features.
Note this networks does not allow dymanic addition/removal of a target.

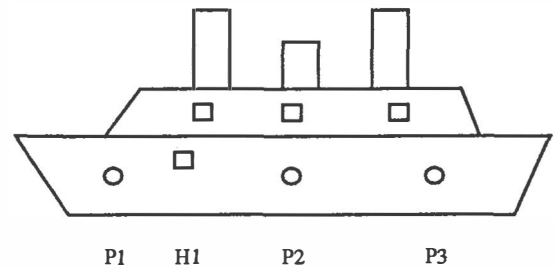

Possible Interpretations for 2 lights:
{P1,P2}{P1,P3}{P2,P3}{P1,H1}{H1,P2}{H1,P3}

Figure 4: This figure illustrates the exhaustive porthole solution. The target is assumed to have 3 portholes (P1, P2,P3) which may be illuminated. In addition to the porthole, this target also has an open hatch (H1) which may appear like a porthole from a distance.

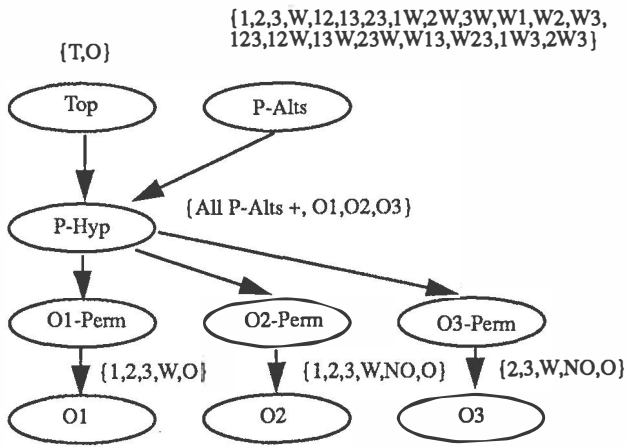

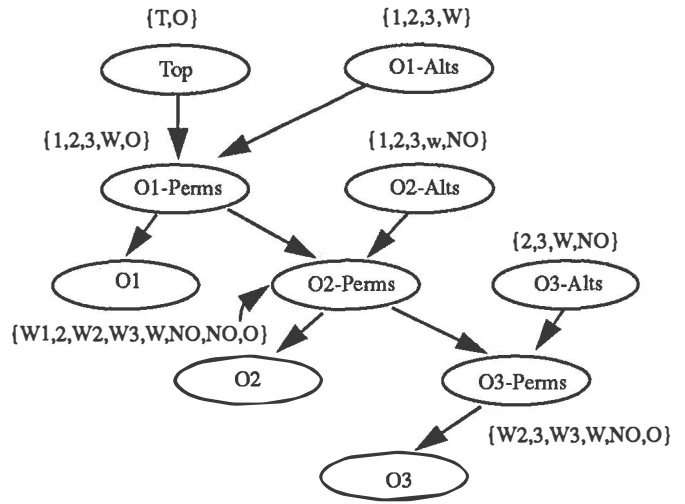

For O1 the hypotheses are: {10%,20%,30% ..... 90%}
For O2,O3 the hypotheses are: {10%,20%, 30%, ......90%, NOT-OBS}

**Figure 5:** This figure illustrates the exhautive-3W Network which allows 3 observations and assumes the possibility of a single incorrect observation.

**Figure 7:** This figure illustrates the SD network for the 3W problem.

If we observe a porthole which appears to be located about 20% of the length of the Target, our evidence might be the following likelihood ratios.
{5:20:5:1:1:1:1:1:1:1}
The resultant Bel* value for this evidence would be:
Bel*(T) = 0.778
Bel*(O) = 0.222

If we observe a porthole which appears to be located about 20% of the length of the Target, our evidence might be the following likelihood ratios.
{5:20:5:1:1:1:1:1:1:1}
The resultant Bel* value for this evidence would be:
Bel*(T) = 0.778
Bel*(O) = 0.222

**Figire 6:** An example Bel* value obtained from adding evidence to the exhaustive-3w network shown in Figure 5.

**Figure 8:** Bel* value as evaluated from the SD-3W network in Figure 7 when given the same evidence as in Figure 6.